# Leveraging Generative AI to Enhance Synthea Module Development

**Mark A. Kramer*, Aanchal Mathur, Caroline E. Adams, Jason A. Walonoski**

**MITRE Corporation, 202 Burlington Street, Bedford, MA**

* **Corresponding author**

## NOTICE

# Abstract

This paper explores the use of large language models (LLMs) to assist in the development of new disease modules for Synthea, an open-source synthetic health data generator. Incorporating LLMs into the module development process has the potential to reduce development time, reduce required expertise, expand model diversity, and improve the overall quality of synthetic patient data. We demonstrate four ways that LLMs can support Synthea module creation: generating a disease profile, generating a disease module from a disease profile, evaluating an existing Synthea module, and refining an existing module. We introduce the concept of *progressive refinement*, which involves iteratively evaluating the LLM-generated module by checking its syntactic correctness and clinical accuracy, and then using that information to modify the module. While the use of LLMs in this context shows promise, we also acknowledge the challenges and limitations, such as the need for human oversight, the importance of rigorous testing and validation, and the potential for inaccuracies in LLM-generated content. The paper concludes with recommendations for future research and development to fully realize the potential of LLM-aided synthetic data creation.

# Conflicts of Interest

The authors declare that they have no financial, commercial, or proprietary interest in Synthea, which is an open-source software tool. The authors are employed by MITRE Corporation, which initiated and maintains Synthea as a non-profit public interest project; none receives direct financial benefit from its use or adoption. MITRE Corporation is a not-for-profit organization that operates multiple Federally-Funded Research and Development Centers. This research was supported by the Centers for Medicare & Medicaid Services (CMS) Office of Information Technology (OIT). The authors have no competing interests to declare.

# Introduction

Synthea (Walonoski et al., 2018) is an open-source software tool for generating synthetic patient data. It creates artificial patient health records that span lifetimes, designed to be statistically realistic while not representing any actual individuals. Synthea has been applied to healthcare research, education, and software development. Synthea operates by generating a population of patients based on real demographic data and then evolving the history of these patients using a set of modules representing various diseases and care protocols. In Synthea, a module represents a specific disease, condition, or clinical guideline that the synthetic patient generator uses to simulate patient experiences and generate realistic health records. A module is comprised of:

- States, representing a point in the progression of a disease or a patient's health status.
- Transitions, defining the movement between states based on certain conditions, such as age, gender, or the presence of specific symptoms.
- Distributions, used to model the variability in patient attributes, such as the age of onset for a disease or the duration spent in a particular state, or probability of an event.
- Codes, representing diagnoses, procedures, and medications, typically based on standardized healthcare vocabularies like SNOMED CT, ICD-10, and RxNorm.
- Lab results or other observed values, typically paired with LOINC codes.

These modules are written in JSON format. When running a simulation, Synthea uses its collection of modules to generate a patient's journey through various health states, generating healthcare events such as onset of conditions, healthcare encounters, diagnoses, procedures, and medications. The final output is a set of synthetic patient records that include a variety of health data points across the simulated lifetime in formats that comply with healthcare data standards, for example, HL7® FHIR® (Fast Healthcare Interoperability Resources). Synthea's methodology allows for the creation of diverse patient populations with complex medical histories, while ensuring that the data remains artificial and does not compromise real patient privacy. The modular design allows for flexibility in defining new diseases, updating existing modules, and customizing the simulation to specific use cases or populations. Existing modules include cardiovascular disease, diabetes, asthma, and many others.

The Centers for Medicare & Medicaid Services (CMS) has provided substantial support for Synthea through several key initiatives, including this LLM research, to support the simulation of Medicare Fee-for-Service claims that accurately reflect eligible beneficiary characteristics. CMS's backing of Synthea stems from critical needs in broadening data availability while protecting patient privacy, enabling research without confidentiality risks, and supporting health interoperability efforts. Their support demonstrates CMS's commitment to fostering healthcare innovation while maintaining strict privacy standards.

Currently, Synthea module design relies on co-development by clinicians and Synthea experts. While this approach ensures understandability, clinical suitability, and reliability, it can be time-consuming and may result in limited treatment variations. By incorporating LLMs into the module development process, we plan to expand Synthea's module library, increase model diversity, and improve the overall quality and variety of synthetic patient data.

Through a series of experiments, we demonstrate the feasibility of using LLMs for various aspects of synthetic clinical data generation module development. The report includes a discussion of the best practices for prompting LLMs to perform Synthea-related tasks effectively.

# Prior Work

The generation of synthetic healthcare data has been explored to address privacy concerns and the need for large datasets in medical research. While Synthea has been a significant contributor in this domain, other AI-based approaches have also been explored for generating synthetic health records.

Generative Adversarial Networks (GANs) have emerged as a leading method for generating synthetic health data. Choi et al. (2017) introduced medGAN, one of the first applications of GANs to generate synthetic electronic health records (EHRs). This approach demonstrated the ability to preserve statistical properties of the original data while maintaining privacy. Yale et al. (2020) developed CorGAN, an improved GAN architecture specifically designed for generating synthetic EHR data with correlated discrete variables, addressing limitations in earlier models. Baowaly et al. (2019) proposed a conditional GAN model for generating synthetic time series EHR data, showing promise in preserving temporal relationships in patient records.

Variational Autoencoders (VAEs) have also shown effectiveness in synthetic health data generation. Dash et al. (2019) introduced a VAE-based model for generating synthetic EHR data, demonstrating improved performance in maintaining statistical properties compared to earlier GAN-based approaches. Nikolentzos et al. (2023) introduced synthetic health data based on variational graph autoencoders. Expanding the application of VAEs, Biswal et al. (2020) proposed a model for generating synthetic electroencephalogram (EEG) data, showcasing the potential of this approach in neurological applications.

Recent work has involved combining multiple techniques and using transformer architectures. Goncalves et al. (2020) introduced a hybrid GAN-VAE model for generating synthetic EHR data, leveraging the strengths of both architectures to improve data quality and diversity. Chen et al. (2021) developed a transformer-based model for generating synthetic clinical notes, demonstrating the potential of natural language processing techniques in this domain. Addressing privacy concerns in collaborative research settings, Xiao et al. (2022) proposed a federated learning approach for generating synthetic health data across multiple institutions.

Researchers have also focused on developing robust methods for evaluating synthetic data quality. Esteban et al. (2017) introduced a framework for assessing the quality of synthetic health data, considering factors such as statistical similarity, privacy preservation, and utility for downstream tasks. Jordon et al. (2018) developed PATE-GAN, combining differential privacy techniques with GANs to generate synthetic health data with provable privacy guarantees. These evaluation and validation efforts are crucial for ensuring the reliability and usefulness of synthetic health data in research and clinical applications.

The above techniques have limitations. GANs, VAEs, statistical methods, and deep-learning techniques emulate their training data and are limited to the patient population and medical practices represented by that data. They cannot be used to conduct “what if” and counterfactual experiments. Synthea, on the other hand, provides a risk-free environment for evaluating the impacts of changes in care delivery and health policy, for example, exploring how health insurance coverage impacts breast cancer survival rates (Scalfani and Bhada, 2020) or comparing the cost-effectiveness of acute myeloid leukemia treatments (Meeker et al., 2022) . This allows for robust testing and development of healthcare tools and innovations, especially in scenarios where real patient data is scarce or too sensitive to use.

The approach of automating the development of care pathways is not entirely new. Process mining techniques, as explored by Rojas et al. (2016), have been used to derive clinical pathways from real-world data. These techniques involve extracting and analyzing event logs from healthcare systems to model patient journeys and care processes. A review of healthcare process mining is presented by Guzzo et al. (2022). However, our review of the literature found no evidence of LLMs being used for this purpose.

# Out-of-the-Box LLM Capabilities

Since there are no known LLMs specifically trained to produce Synthea modules, the options were either to (1) train an LLM from scratch,  (2) fine-tune an existing LLM, or (3) use a pre-trained general-purpose LLM model without modification. Synthea has fewer than 90 validated modules, far less than needed for training from scratch but conceivably useful for fine-tuning. However, there is a problem creating input-output pairs needed for fine-tuning. The output is known (it is the module itself) and the input should somehow capture the clinical knowledge used to create the module. While some of that knowledge exists in comments and users guides, in most cases the clinical knowledge used to create and validate the module was not recorded comprehensively or systematically at the time of module creation.[1]

Given the remaining option of using pre-trained LLMs, we tested the out-of-the-box capability of three LLMs available in mid-2024, Claude Sonnet 3.5, GPT-4o, and Gemini 1.5 Pro. To establish a baseline, we began with the following bare-bones prompt: “*Please produce a Synthea module for hyperthyroidism. Return the result as JSON.”* Hyperthyroidism was chosen because it is a common disease with readily-available information, but no current Synthea module. Even with no supporting information, two of three LLMs were able to create simplistic Synthea modules following the correct sequence of disease onset, testing, diagnosis, and treatment (Figure 1). However, the modules contained structural and logical problems:

- Failure to capture real-world clinical details (All)
- No delay state or guard after the initial state (causing 100% of the population to contract the disease immediately upon birth) (GPT 4o)
- Syntax errors resulting in no transitions between states (Gemini)
- Unreachable states (GPT 4o)
- Incorrect RxNorm and SNOMED CT codes (All)

To get more satisfactory results from general-purpose LLMs, a more sophisticated approach was needed.

[1] Potential approaches to training, not pursued in this study, involve deriving missing inputs from literature sources or by creating textual summaries of the existing validated modules.

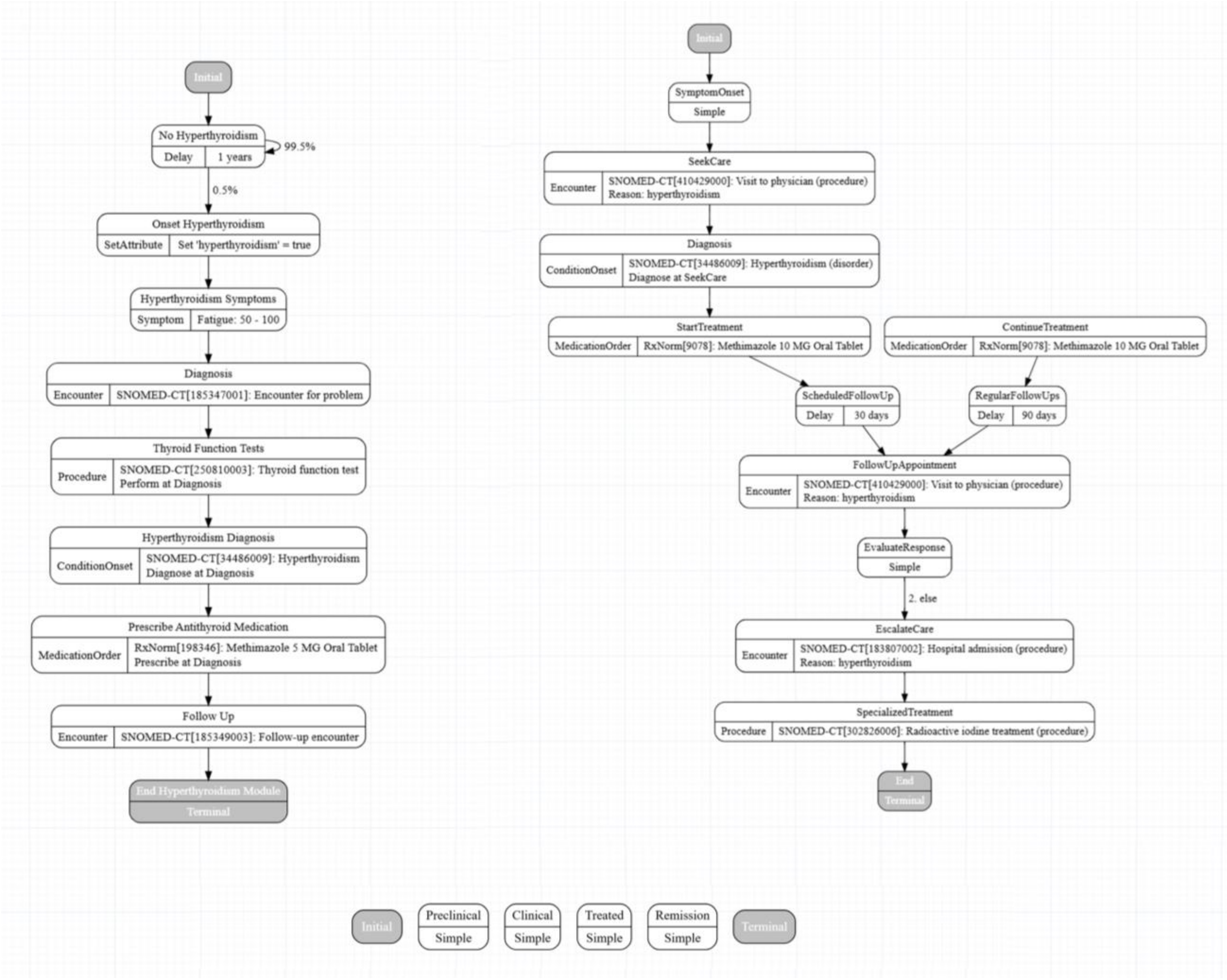

*Figure 1. Synthea modules produced with minimal prompting by Claude 3.5 Sonnet (claude-3-5-sonnet-20240620) (left), GPT-4o (gpt-4o-2024-05-13) (right), and Gemini 1.5 Pro (bottom). Modules are rendered graphically by the Synthea Generic Module Builder (https://synthetichealth.github.io/module-builder/).*

# Methodology

We developed a four-stage methodology for creating Synthea modules using LLMs:

1. Generating disease information in a useful form,
2. Creating an initial disease module from that information,
3. Validating the module, and
4. Iteratively improving the module based on validation results.

While we present these steps as a fully-automated and autonomous process, human input can be inserted at any step of this process, for example, to add or modify information about the disease and its treatments, to manually modify a disease module, or to direct the LLM to make certain changes to the module.

## Disease Profiles

The first stage of the methodology involves gathering and organizing information about the target disease in a form useful for creating a Synthea module. A *disease profile* is defined as a set of statements that describes the key characteristics and management of a specific medical condition. When subsequently we use the word *requirement*, we are referring to a single identifiable (numbered) statement in a disease profile.

Several types of information are needed to produce a robust Synthea module, including, but not limited to:

1. Incidence and prevalence: Statistics on how common the disease is in different populations.
2. Risk factors and demographics: Data on age, gender, race, socioeconomic status, and other factors that influence disease onset and progression.
3. Lifestyle factors: How lifestyle choices (e.g., diet, exercise, smoking) impact disease onset, progression, and management.
4. Screening recommendations: Information on any recommended screening practices for early detection.
5. Disease progression: Information on how the disease naturally progresses over time, including different stages or severity levels.
6. Symptoms: The indicators associated with the disease, including their probability and severity.
7. Diagnostic criteria: Information on how the disease is diagnosed, including specific tests, observations, or criteria used.
8. Treatment options: Details on various interventions, medications, and procedures used to treat the disease.
9. Treatment outcomes: Data on how different treatments affect disease progression, symptom relief, and overall patient health.
10. Follow-up care: Details on typical follow-up procedures, monitoring, and long-term management of the disease.
11. Complications: Information on potential complications or related conditions that may develop.
12. Prognosis: Long-term outlook for patients with the disease, including potential for recovery or disease progression.

The ideal way to create a disease profile is to engage medical experts to compile and organize relevant medical information from authoritative sources such as clinical guidelines, research papers, and expert knowledge. However, this process can be challenging and time consuming since people with clinical and epidemiologic expertise typically lack Synthea expertise, and vice versa.

After some experimentation, good results were obtained by supplementing the built-in knowledge of the LLMs with hand-selected sources. For Synthea purposes, the most relevant information emphasizes disease progression and treatment pathways aligned with current standards of care. Using this information as context, the LLM is prompted to elicit the types of medical knowledge listed above. Because Synthea uses quantitative information to generate realistic populations, the prompt (given in Appendix A.1 Prompt for Disease Profile Generation) emphasizes the importance of including quantitative data such as percentages, probabilities, and thresholds. The prompt also includes an example of a disease profile to demonstrate the desired output format and proper level of detail.

In the cases we studied, the LLM-generated disease profiles required expert validation and manual adjustment, particularly for quantitative metrics. Incidence rates for various subpopulations can be complicated to compute, and information in literature is often incomplete and contradictory. The source citations elicited in the prompt and supplied by the LLM help human reviewers validate the LLM results.

Using an explicit disease profile anchors the Synthea module in medical reality and enables the LLM to focus solely on bridging the gap between medical knowledge and the structured format required for synthetic patient data generation. The disease profile also serves as a reference point for reviewing the generated module, enabling iterative refinement and improvement of the module in subsequent steps of the module generation process.

## Initial Module Generation

The second stage involves generating an initial module, using the disease profile as input. To do this, we created a structured prompt containing the following parts:

1. Background: A short introduction to Synthea, including rules that must be followed to create valid Synthea modules.
2. Synthea Reference: Technical documentation describing types of Synthea states and transitions.
3. Synthea Examples: Another document containing several examples of Synthea modules in JSON format, taken from the existing module set.
4. Disease profile: The disease profile generated by the previous step.
5. Task: A description of the LLM's assignment, namely, to create a Synthea module based on the disease profile, with additional details on how the LLM should document the relationship between requirements in the disease profile and states in the Synthea module, and a description of the desired output format.

As part of the Task, the LLM was instructed to add a property to each Synthea state that captures the requirement number(s) that justify the state, providing traceability between the disease profile and the module. The full prompt is given in Appendix A.2 Prompt for Initial Module Generation.

# Model Validation

Validation, the third stage of the method, is crucial for ensuring the generation of high-quality synthetic patient data that accurately represents real-world health scenarios. Our method uses a two-level validation technique to assess and improve the quality of Synthea modules. Level 1 validation focuses on verifying that the module adheres to the basic structural requirements of Synthea, including valid JSON formatting and compliance with essential module rules. Level 2 involves validating the module's states, transitions, and parameters against the disease profile. This two-tiered approach ensures not only the technical correctness of the module (e.g., it functions as software) but also its clinical fidelity relative to stated requirements of the disease profile. Level 3 (not implemented) validates the population, incidence, and prevalence associated with a disease process.

## Level 1: Structural and Syntax JSON Validation

The first level of validation for Synthea modules focuses on ensuring structural integrity and adherence to Synthea guidelines. Because of the way it represents states, Synthea JSON does not lend itself to schema validation. Customized unit tests are therefore required to check the overall structure and adherence to guidelines in state definitions, transitions, logic, and other elements. In the following, we make use of terms and concepts defined in the Synthea documentation.[2] In brief, each module has a set of states and transitions. A state represents a health or demographic status or clinical activity. Every module begins with an Initial state and most modules end with a Terminal state. Common states include Encounter, ConditionOnset, Symptom, Procedure, and MedicationOrder. States also include control actions, such as delays, which allow a certain amount of time to pass, and guards, which pause until a certain condition is met. Transitions represent movement between states, for example, moving from a healthy to unhealthy state, or receiving a treatment. Transitions come in several varieties, the most common being Direct (transition to a single state), Distributed (probabilistic transition to one of several possible states), and Conditional (transition to one of several states by conditional logic). Modules can invoke submodules, reusable sets of states that encapsulate common medical procedures.

Key validation checks include:

- Path integrity: Ensuring a direct path exists from the Initial state to every other state, maintaining a continuous progression through the module.

[2] https://github.com/synthetichealth/synthea/wiki

- State, attribute, and submodule validity: Checking if all submodules invoked by the module exist, all states called in the module are initialized before they are referenced, and all attributes referenced in the module are assigned earlier in the module or are valid attributes of other modules.[3]
- Transition completeness: Verifying that every state, except the Terminal state, has an output transition, preventing dead-ends in patient trajectories.
- Temporal logic: Confirming the presence of a guard or delay immediately following the Initial state, simulating realistic start conditions and time progression with respect to disease onset and healthcare interactions.
- Clinical workflow accuracy: Validating that each Encounter state is paired with a subsequent EncounterEnd state, mirroring real-world clinical episodes.
- Care delivery sequence: Ensuring all clinical actions (e.g., Procedure, Observation, MedicationOrder, ImagingStudy) occur between Encounter and EncounterEnd states, reflecting proper care delivery workflows.
- Event timing: Confirming that ConditionOnset, Symptom, and ConditionEnd states reflect proper ordering of condition progression and conclusion.
- Probabilistic integrity: Verifying that weights in distributed transitions sum to 100%, ensuring accurate representation of branching probabilities.

This automated validation process generates a set of error messages for rule violations and warnings for potential issues. As discussed below, these issues are fed back to the LLM, ensuring that the modules maintain a basic level of integrity, without delving into conformance to the disease profile.

## Level 2: Clinical Accuracy Validation

The next level focuses on ensuring the clinical accuracy of the Synthea module by comparing it against the disease profile. The disease profile, described above, is a compendium of statements about the disease being modeled, taken as the ground truth for clinical accuracy. The review process involves using an LLM to carry out the following tasks for each requirement in the disease profile:

1. Determine how well the requirement is implemented in the Synthea module.
2. Create suggestions for changes or improvements to the implementation, if necessary.
3. Assign a score (0 to 1.0) to indicate the quality of implementation for each requirement.

The scoring system is as follows:

- 0.0: Requirement not implemented
- 0.25: Implemented incorrectly
- 0.50: Partially implemented
- 0.75: Fully implemented but with minor issues
- 1.00: Fully and correctly implemented

Subjectivity in the scoring rubric is unavoidable because there is no general, algorithmic way to calculate how much of a requirement has been implemented. We allow the LLM to interpret the scoring system but require the LLM to provide an explanation of why the score was chosen, allowing us to judge the reasonableness of the scores. The prompt implementing the Level 2 review is given in Appendix A.3 Prompt for Clinical (Level 2) Review.

## Level 3: Population Validation

A third level of module validation, not considered in this study, involves evaluating the population produced by running the model and determining whether that population aligns with expected demographic and clinical distributions. Statistical validation of synthetic populations has been studied by Goncalves et al. (2020) and Chen et al. (2021). Key factors to consider include disease incidence and prevalence in various population groups, the proportion of patients receiving various treatments, and the rates of key events such as remission, relapse, and mortality.

---

[3] Attributes are read-write variables on patients. Attributes are accessible system-wide and can be used by more than one module. When they are used by more than one module, this represents and explicitly defines a dependency between these diseases.

To perform meaningful population validation, there must be a synthetic population large enough to obtain many samples of patients with the target condition. The required population size varies significantly based on condition prevalence. In this study, we did not include population-level validation as part of the iterative generation process because of the time required to generate a large synthetic population on each iteration. In the future, we envision using Level 3 validation outside of the iterative process to set parameters once the structural elements and flows of the module have been established.

## Progressive Refinement Via Iterative Generation

The fourth major stage of the process employs a feedback loop that leverages validation results to guide subsequent improvements in the Synthea module. This process has several steps:

1. Level 1 review and regeneration: After module generation, a Level 1 review is conducted, and errors are collected in a narrative form suitable for inclusion in a prompt instructing the LLM to correct its errors. The prompt includes a list of errors, suggestions to correct the errors generated by the LLM, and the module to correct.
2. Level 2 review: After one opportunity to correct Level 1 errors, an LLM performs the Level 2 review. The review summarizes how each requirement is or is not implemented in the current module, how the module needs to change to better represent the requirement and provides a score in the range 0 to 1.0 in 0.25 increments.
3. Targeted Requirements: The system next identifies requirements with the lowest scores and selects a limited number of these to be the focus for the next iteration. This approach allows for targeted improvement efforts, focusing on requirements most in need of improvement. For each low-scoring requirement, the system compiles the feedback provided by the LLM, including the requirement number, the current implementation status, and the LLM's recommendations for improvement. This results in a prompt that is fed back into the module generation process.
4. Continuous Validation: After each regeneration step, the new module undergoes another round of review and scoring. This re-validation ensures that improvements in one area do not inadvertently introduce issues in others and allows for tracking of the module's overall progress. The process continues iteratively until a satisfactory average score across all requirements is achieved or until the improvement between iterations becomes negligible.

Several aspects are worth noting:

1. Optimization Objective: The iterative process maximizes the average Level 2 score, which gives equal weight to each requirement. It would be trivial to prioritize certain requirements more than others by implementing weighted averages.
2. Improvement Trajectories: Initial iterations may see rapid improvements as major issues are addressed, followed by a plateau as more nuanced refinements are needed. Monotonic improvement is not guaranteed, since improvements in one requirement may affect the implementation of others due to the interconnected nature of clinical processes. However, non-productive steps can be rejected by returning to the previous iteration.
3. Stochasticity: Because LLM outputs have a degree of randomness, there is no guarantee of convergence to a unique structure. Running multiple refinement sequences with different initial generations could help assess the robustness of the final module and allow for human review of choices for potential feasible solutions.

The prompt for the iterative refinement process is given in Appendix A.4 Prompt for Iterative Improvement.

# Implementation

The module generation algorithm was implemented in Python, using API interfaces to Claude 3.5 Sonnet, GPT-4o, and Gemini 1.5 Pro. The use of proprietary, closed source models reflected our discovery that only the most powerful models were able to generate answers of sufficient detail to be useful for Synthea module generation..

## Validating Module Review and Scoring

We used an LLM to evaluate how well each requirement from the disease profile was represented in the module. The LLM can be the same or different than the one used to generate the module, but to assure an objective result, the review took place in an independent session, i.e., one that does not share history with the generation step.

Prior to full-scale implementation, we ran a series of tests to assure that the Level 2 (clinical) reviews could be relied upon to reflect the quality of the module, both in terms of accuracy and precision. To evaluate precision, we generated review ratings using three different Synthea modules and three different LLMs, with 5 runs for each combination. The three modules used in this test were all versions of hyperthyroidism produced by Gemini, GPT, and Claude, respectively. Overall, Claude showed the best precision (lowest standard deviation), and GPT the worst (Table 1). GPT also gave the highest average scores, while there was a high level of agreement between the scores given by Gemini and Claude. We therefore selected Claude as the reviewer for all subsequent experiments.

*Table 1. Scores (as percentages) from Level 2 reviews by GPT-4o (gpt-4o-2024-08-06), Claude 3.5 Sonnet (claude-3-5-sonnet-20240620), and Gemini 1.5 Pro. Each module-LLM combination was tested 5 times to gauge reproducibility.*

| | Module 1 | | | Module 2 | | | Module 3 | | |
|---|---|---|---|---|---|---|---|---|---|
| **Reviewer** | GPT | Claude | Gemini | GPT | Claude | Gemini | GPT | Claude | Gemini |
| Run 1 | 30.1 | 39.8 | 35.8 | 69.6 | 51.1 | 65.5 | 100.0 | 77.8 | 81.8 |
| Run 2 | 39.2 | 36.1 | 42.3 | 93.2 | 51.7 | 52.8 | 100.0 | 65.3 | 80.4 |
| Run 3 | 39.8 | 40.3 | 40.3 | 69.1 | 53.3 | 45.3 | 93.2 | 74.4 | 73.4 |
| Run 4 | 52.3 | 35.8 | 30.1 | 51.4 | 53.3 | 51.2 | 79.6 | 76.1 | 68.6 |
| Run 5 | 34.7 | 28.4 | 38.1 | 95.0 | 55.0 | 48.0 | 85.2 | 78.4 | 69.2 |
| **AVERAGE** | **39.2** | **36.1** | **37.3** | **75.7** | **52.9** | **52.6** | **91.6** | **74.4** | **74.7** |
| **STDEV.S** | **8.3** | **4.8** | **4.7** | **18.4** | **1.5** | **7.8** | **9.1** | **5.3** | **6.2** |

## Input and Output

Our approach required about 50K input tokens for the background, reference manual, and examples. This fit comfortably within the 200K context window of Claude 3.5 Sonnet, the 128K window of GPT-4o, and Gemini's 1M input window. Unfortunately, modules sometimes exceeded the maximum output length of 4,096 tokens for GPT-4o and 8,192 for Gemini 1.5 Pro and Claude 3.5 Sonnet, causing the LLM to abruptly truncate its response, often mid-word, mid-number, or within a critical JSON structure. The system automatically requested continuations, but simply concatenating successive outputs often resulted in invalid JSON due to overlaps, discontinuities, and other syntactic issues at juncture points. We developed specialized functions to splice continuation outputs to ensure the resulting JSON was both valid and complete. This approach allowed for the generation of large, complex modules exceeding the current output limits of the LLMs.

## Feedback Strategy

A crucial component of our implementation was the integration of Level 1 and 2 review feedback into the iterative improvement cycle. As touched on previously, we batched 10 problems at a time, prioritizing the most critical issues, for inclusion in the iterative improvement prompt. This approach allowed the LLM to concentrate on key areas for improvement without being overwhelmed by excessive feedback. Additionally, we implemented a randomization strategy for the selection and order of individual requirements within each batch. This ensured that equal attention was given to issues regardless of their position in the disease profile, preventing bias towards requirements earlier in the list. As the lowest scores were addressed, attention would automatically shift to higher-scoring requirements, thus addressing increasingly nuanced aspects of the module.

# Example

Hyperthyroidism (HT) was chosen as an example for several reasons. First, Synthea currently lacks a module for HT, presenting an opportunity to fill a gap in its capabilities. Second, treatment and management of HT is complex, which showcases the potential benefits of our automated approach. Third, the disease and its treatments are relatively standardized and well-documented, which allowed for objective evaluation of the result.

The final modules are too lengthy for inclusion in this paper but can be found at: https://github.com/synthetichealth/synthea-llm/tree/main/modules.

## Disease Profile Generation

The first stage involved prompting Claude to create a disease profile for hyperthyroidism. Initially we relied on the LLM's internal knowledge of hyperthyroidism, which resulted in a disease profile with 36 requirements (see Appendix B.1 Disease Profile Based on Claude Sonnet Intrinsic Knowledge).

We shared the automatically-generated disease profile with several physicians employed by MITRE. Most felt that the profile covered all major points in at least a superficial manner. However, we received feedback that the disease profile was simplistic and failed to capture the complexity of treating this disease. Another pointed out the lack of citations in the profile. One MD viewed the probabilities and percentages as mostly plausible, but felt that some were misestimated, perhaps insertions from papers that measured sensitivity and specificity in one context but reapplied out of context.

Based on that experience, we tried again, this time including validated disease guidelines in the input to the LLM. These sources are listed in Appendix B.2 Curated Knowledge Sources for Hyperthyroidism Disease Profile. This resulted in a more accurate and detailed list of requirements. However, the incidence of disease among different subpopulations, as required by Synthea, was not clearly expressed and so the incidence rates were calculated manually. In addition, we added several subsections to the disease profile in addition to the requirements: general background on hyperthyroidism, assumptions, and a list of acronyms. The resulting profile is shown in Appendix B.3 Disease Profile With Curated Knowledge and Clinical Expertise. This was the basis for subsequent experiments.

## Initial Module Generation

With the disease profile and the prompt in Appendix B.3 Disease Profile With Curated Knowledge and Clinical Expertise, initial Synthea modules were generated using GPT, Claude, and Gemini. In the initial generation, prior to any progressive improvement, there was a wide range quality, measured by Level 1 warnings and Level 2 scores, and sophistication, measured by the number of states. A summary of results over the 15 initial runs is given in Table

2. While GPT produced valid JSON in each of 5 runs, it also had the lowest Level 2 scores and produced modules with the fewest states. Gemini created modules with the most states and the highest Level 2 scores, but also produced invalid JSON in 2 of 5 runs. An example of one of the initial modules is shown in Figure 2.

*Table 2. Results of initial module generation with Level 2 scores, number of states, and number of warnings given by the Synthea Module Builder.*

| | **GPT** | | | | **Claude** | | | | **Gemini** | | | |
|---|---|---|---|---|---|---|---|---|---|---|---|---|
| | Valid JSON | L2 Score | # States | Warnings | Valid JSON | L2 Score | # States | Warnings | Valid JSON | L2 Score | # States | Warnings |
| Run 1 | Yes | 38.6 | 43 | 7 | No | ? | ? | ? | Yes | 77.8 | 73 | 1 |
| Run 2 | Yes | 50.0 | 22 | 1 | Yes | 74.4 | 82 | 0 | No | ? | ? | ? |
| Run 3 | **Yes** | **45.6** | **30** | **0** | **Yes** | **63.1** | **72** | **0** | No | ? | ? | ? |
| Run 4 | Yes | 34.4 | 31 | 0 | Yes | 56.1 | 64 | 29 | Yes | 97.7 | 101 | 5 |
| Run 5 | Yes | 21.1 | 25 | 14 | Yes | 63.9 | 101 | 0 | **Yes** | **73.9** | **95** | **0** |
| **Average** | **100%** | **37.9** | **30.2** | **4.4** | **80%** | **64.4** | **79.8** | **7.3** | **60%** | **83.1** | **89.7** | **2.0** |

We employed a strategy of creating multiple initial modules and selected the most promising modules for further development. We selected one run for each LLM as starting points for iterative improvement based on fewest L1 warnings and secondarily, highest L2 score. For runs with similar L1 and L2 scores, we selected the modules that achieved the score parsimoniously, i.e., with the fewest states: specifically, GPT Run 3, Claude Run 3, and Gemini Run 5.

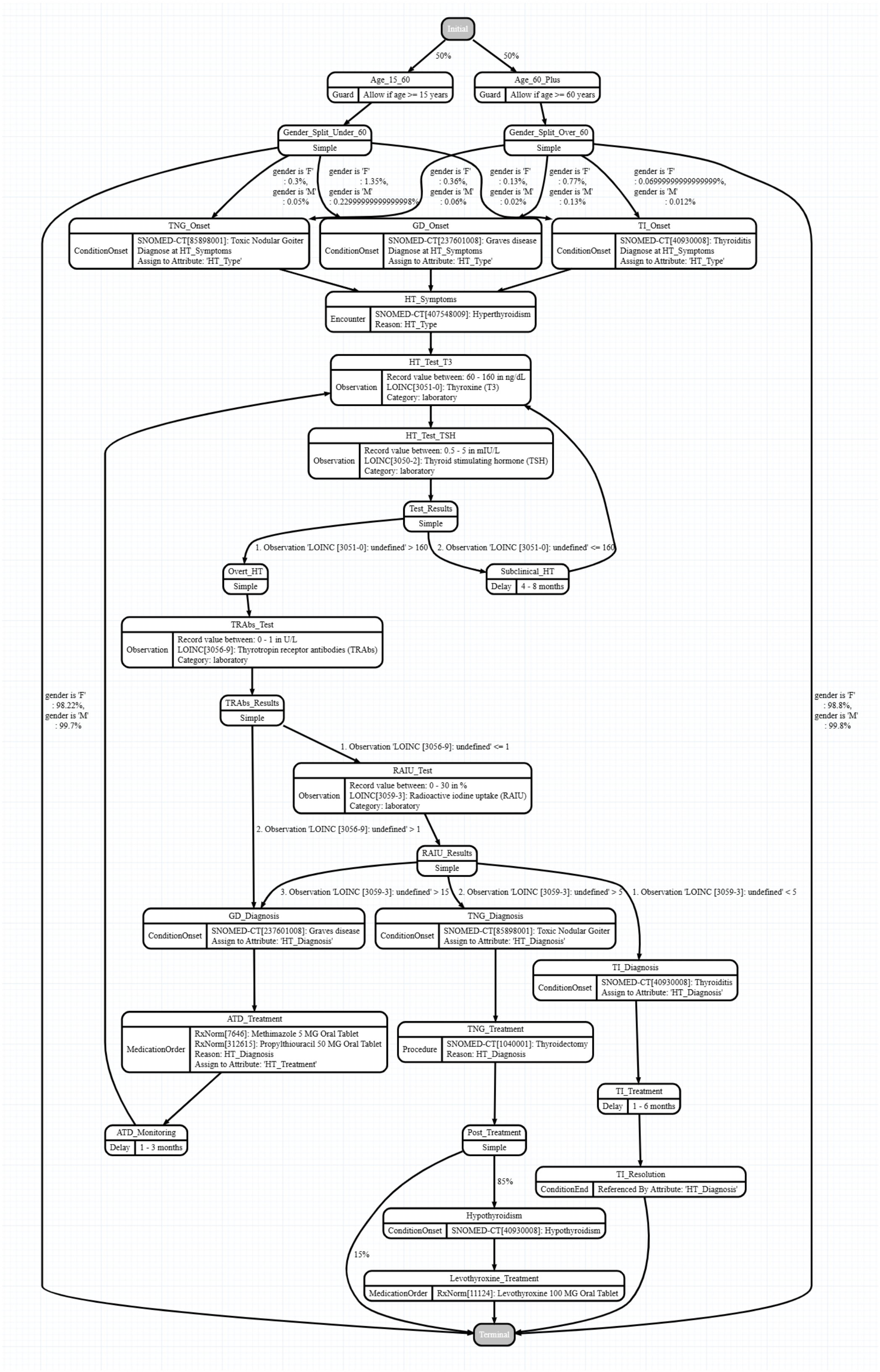

*Figure 2. Initial module for hyperthyroidism generated by GPT-4o.*

# Iterative Refinement

The process of iterative refinement was begun after reviewing the first generated module. An excerpt from a typical Level 2 review is shown in Table 3. As discussed previously, the lowest-scoring requirements and the suggested changes were incorporated into the prompt that led to the next iteration. This process was repeated 10 times. Figure 3 shows the progress of the overall Level 2 score as a function of the iteration.

*Table 3. Excerpt from Level 2 review of initial module produced by GPT-4o.*

| Req.Number | Requirement | Explanation | Transitions | Change | Score |
|---|---|---|---|---|---|
| **23** | If initial or follow-up testing reveals overt HT, and palpable thyroid nodules are present, or physiologic signs of GD are unclear, then TRAbs should be measured next. | The module partially implements this requirement. After the Test_Results state determines overt HT, it transitions to the TRAbs_Test state. However, it does not consider the presence of palpable thyroid nodules or unclear physiologic signs of GD. | The transition from Test_Results to TRAbs_Test for overt HT cases is correct, but it lacks the conditional elements specified in the requirement. | Modify the Test_Results state to include checks for palpable thyroid nodules and unclear GD signs before transitioning to TRAbs_Test. | 0.75 |
| **24** | However, if initial or follow-up testing reveals overt HT and there are no palpable thyroid nodules and there are clear physiologic signs of GD, TRAbs testing is not necessary, and a diagnosis of GD is confirmed. | The module does not implement this requirement. It always proceeds to TRAbs testing for overt HT cases without considering the absence of thyroid nodules or presence of clear GD signs. | The transition from Test_Results to TRAbs_Test does not account for this scenario. | Modify the Test_Results state to include a direct transition to GD_Diagnosis when overt HT is present, there are no palpable thyroid nodules, and there are clear GD signs. | 0.00 |
| **25** | If TRAbs are elevated, a GD diagnosis is confirmed. TRAbs will be elevated in 98% of cases of GD. | The module partially implements this requirement. The TRAbs_Results state transitions to GD_Diagnosis if TRAbs are elevated. However, it does not accurately reflect the 98% probability of elevated TRAbs in GD cases. | The transition from TRAbs_Results to GD_Diagnosis when TRAbs are elevated is correct, but the probability does not match the requirement. | Adjust the TRAbs_Test state to ensure that 98% of GD cases have elevated TRAbs. This may require restructuring the module to determine the underlying condition before the test. | 0.75 |
| **26** | Alternatively, if TRAbs are normal, radioactive iodine uptake test (RAIU) should be conducted. However, RAIU is contraindicated in pregnancy and lactation, and a thyroid ultrasound with color-flow Doppler procedure should be substituted. | The module partially implements this requirement. If TRAbs are normal, it proceeds to the RAIU_Test. However, it does not consider contraindications for RAIU or provide an alternative thyroid ultrasound option. | The transition from TRAbs_Results to RAIU_Test when TRAbs are normal is correct, but it lacks the conditional elements for pregnancy and lactation. | Add a check for pregnancy and lactation before the RAIU_Test state. If contraindicated, add a transition to a new Thyroid_Ultrasound state. | 0.50 |
| **27** | If GD is present, RAIU will reveal diffusely increased uptake in 95% of cases, and then GD diagnosis is confirmed. | The module partially implements this requirement. The RAIU_Results state transitions to GD_Diagnosis if RAIU is high. However, it does not accurately reflect the 95% probability of increased uptake in GD cases. | The transition from RAIU_Results to GD_Diagnosis when RAIU is high is correct, but the probability and threshold do not match the requirement. | Adjust the RAIU_Test and RAIU_Results states to ensure that 95% of GD cases have diffusely increased uptake (>15%). | 0.75 |
| **28** | Alternatively, if TNG is present, RAIU will reveal focal areas of increased uptake, and then TNG diagnosis is confirmed. Nodules revealed by ultrasound likewise indicate the presence of TNG. | The module partially implements this requirement. The RAIU_Results state transitions to TNG_Diagnosis if RAIU is in a middle range. However, it does not explicitly model focal areas of increased uptake or consider ultrasound results. | The transition from RAIU_Results to TNG_Diagnosis when RAIU is in a middle range is a simplification of the requirement. | Modify the RAIU_Test and RAIU_Results states to better model focal areas of increased uptake. Add a Thyroid_Ultrasound state that can also lead to TNG_Diagnosis if nodules are present. | 0.50 |
| **29** | Alternatively, if TI is present, RAIU will reveal low or absent uptake, and a diagnosis of TI can be confirmed. | The module implements this requirement. The RAIU_Results state transitions to TI_Diagnosis if RAIU is low. | The transition from RAIU_Results to TI_Diagnosis when RAIU is low (<5%) correctly represents this requirement. | none | 1.00 |

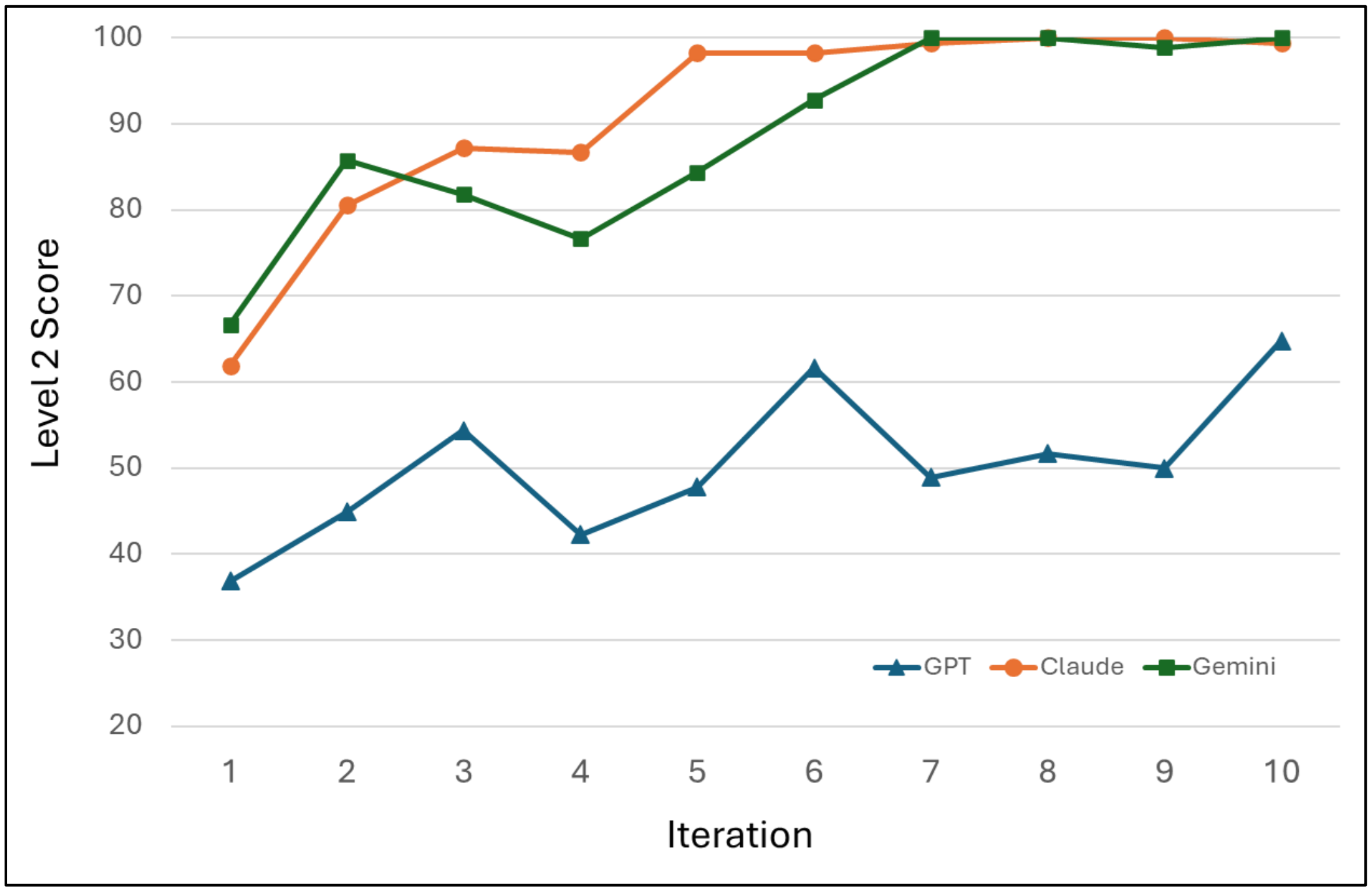

*Figure 3. Overall Level 2 (clinical accuracy) score as a function of iteration.*

To satisfy all the requirements of the disease profile, the complexity of the modules measured by the number of states tended to increase with additional iterations. For example, the module created by Claude started with 72 states and increased to 109 states by the fifth iteration.

## Human Evaluation

To assess the quality of LLM-generated Synthea modules, a Synthea expert provided an in-depth analysis of the modules created by Claude and Gemini at the end of the iterative improvement phase, both of which scored near 100% in Level 2 validation. The expert evaluated successful elements, areas of the module needing improvement, and incorrect implementations in each module.

The evaluations revealed that while the LLMs generated runnable modules, neither produced production-ready code. The Claude-generated module required improvements in areas such as pairing encounter beginnings and endings, duplication of condition onsets, and elimination of unnecessary states. The Gemini-generated module showed a better foundational understanding of symptom probability chains and incorporation of high-risk patient designation within the module. Still, the Gemini module had more structural problems, including missing transitions, incorrect attribute usage, and references to values within observation states that were not defined. Both modules consistently hallucinated formal medical codes.

Overall, the Synthea expert concluded that even with these flaws, the resulting modules exhibited a high level of sophistication in terms of technical implementation detail and in overall construction, design, and organization.

# Discussion

The results demonstrate that LLMs can assist in creating complex Synthea modules through a structured, iterative approach. Several key findings emerge from this research:

First, the progressive refinement methodology proved effective at improving module quality. All three tested LLMs showed significant improvement in Level 2 (clinical accuracy) scores through successive iterations, with final scores approaching 100% for two of the LLMs. This suggests that even when initial generations are flawed, systematic feedback and refinement can produce high-quality results.

Second, the choice of LLM impacts both the initial quality and the refinement trajectory. While GPT-4o produced consistently valid JSON, its initial clinical accuracy scores were lower (average 37.9%) compared to Claude (64.4%) and Gemini (83.1%). However, Gemini and Claude eventually converged to similar high-quality results through iteration, suggesting the methodology can compensate for initial differences in model performance.

Third, the use of curated knowledge sources significantly improves the quality of disease profiles compared to relying solely on the LLMs' internal knowledge. This highlights the importance of combining LLM capabilities with authoritative medical sources rather than treating LLMs as standalone knowledge bases.

Fourth, the automated review process proved reliable and consistent, particularly when using Claude as the reviewer, with standard deviations in generated scores typically below 5%. This enables tracking of improvements across iterations.

The approach demonstrates value in handling complex disease pathways with multiple treatment options and decision points. The hyperthyroidism case study illustrates how LLMs can manage intricate clinical logic while maintaining adherence to both medical accuracy and Synthea's structural requirements.

# Limitations

Several important limitations should be noted:

1. Medical Code Accuracy: While the LLMs can create structurally valid modules with plausible clinical pathways, the codes generated for diagnoses, procedures, and medications are frequently incorrect. A follow-on step to detect and replace hallucinated codes is necessary.

2. Integration Challenges: The current approach treats each disease module in isolation, without considering interactions with other conditions or comorbidities. This limits the realism of the synthetic patient data, as real patients often have multiple interacting conditions that affect treatment decisions and outcomes.

3. Validation Constraints: While Level 1 (structural) and Level 2 (clinical) validations are well-defined, Level 3 (population) validation remains challenging due to limited access to comprehensive real-world data for comparison. This makes it difficult to verify whether the synthetic populations accurately reflect real-world epidemiological patterns. This is an issue that exists broadly in the domain of synthetic data generation and is not exclusively a problem related to the use of LLM workflows.

4. Resource Requirements: The methodology requires non-trivial computational resources. A single end-to-end generation process typically required 20-30 minutes to complete. However, the cost of the commercial LLM services (less than $500 for the entire project) was insignificant compared to the cost of skilled personnel.

5. Clinical Expertise Requirements: While LLMs can assist in module creation, the process still requires significant clinical expertise, particularly in validating disease profiles and reviewing generated modules. This limits the extent to which the process can be fully automated.

6. Negative Rules: The current scoring system for Level 2 does not instruct the LLM on what *not* to do. Because of this omission, we observed certain unrealistic behaviors, for example, encounters lasting for many months over entire courses of treatment.

7. Excessive State Proliferation: Through successive iterations, modules tended to grow and become more complex. While this can reflect real-life patient pathways, it may also indicate opportunities for optimization and simplification.

# Conclusion

This research demonstrates that LLMs can serve as valuable tools in the development of synthetic health data generators when used within a structured, iterative methodology. The *progressive refinement* approach successfully bridges the gap between medical knowledge and synthetic data generation, enabling the creation of complex disease modules that capture realistic clinical pathways.

The methodology's success in creating a hyperthyroidism module suggests potential applicability to a wide range of medical conditions, particularly those with well-documented clinical guidelines and treatment pathways. The approach could significantly accelerate the development of new Synthea modules, helping to expand the variety and sophistication of synthetic patient data available for healthcare research and development.

However, the limitations identified indicate that LLMs should be viewed as assistive tools rather than autonomous module creators. The need for clinical expertise, manual code verification, and careful validation remains critical. Future work should focus on addressing these limitations, particularly in areas such as medical code accuracy, comorbidity handling, and population-level validation.

The methodology introduced here represents a significant step forward in synthetic health data generation, offering a systematic approach to leveraging LLM capabilities while maintaining clinical accuracy and practical utility. As LLM technology continues to evolve, this framework provides a foundation for further advancement in synthetic health data generation, ultimately supporting broader access to realistic but non-identifiable patient data for healthcare innovation.

Finally, this research suggests the potential utility of using LLM workflows more broadly within modeling and simulation, beyond the scope of disease modeling.

# Acknowledgements

The authors wish to thank Karl Davis and Cynthia Miles of CMS OIT for their support and guidance during this work. The work was funded by CMS under Contract ID 75FCMC18D0047, Task Order 75FCMC19F00010.

# Appendix A: LLM Prompts

In the following curly brackets {} indicate places where text substitution takes place.

## Appendix A.1 Prompt for Disease Profile Generation

Please create a comprehensive disease profile for {disease} in the style of the example provided. Use these externally sourced documents about {disease} to provide more accurate medical context to the disease profile: {combined text}.

Additionally, use the information provided in the {image information} as well to understand information about the disease progression of {disease}.

The output of the result should be a numbered list of facts about the disease specifically covering the following aspects, where applicable:

- Prevalence and incidence in the general population and specific subgroups (e.g., by age, gender, ethnicity)
- Risk factors with associated increased risk percentages
- Etiology, including common causes and their relative frequencies
- Pathophysiology
- Symptoms and clinical presentation, including probability of each symptom
- Diagnostic criteria and tests, with sensitivity and specificity data
- Differential diagnoses
- Screening recommendations and prevention strategies, if applicable
- Treatment options including a. First-line treatments b. Alternative treatments c. Success rates for each treatment d. Criteria for selecting different treatment approaches e. Common side effects and their frequencies
- Monitoring and follow-up protocols
- Short-term and long-term outcomes
- Complication rates and types
- Recurrence or progression rates
- Survival rates, if applicable
- Quality of life impacts

For each numbered point in the profile, refer to the specific source of information from the externally sourced documents by including the label that represents where the information for that point came from. The labels in the external content are in the format {{X.Y}}, where X represents the source number and Y represents the line number within that source and occur at the start of each line. Ensure these labels are clearly marked as references and are not part of the content. Multiple points of information can be sourced from the same line.

Do not use vague words or general statements within the disease profile. Please provide quantitative data whenever possible, such as percentages, probabilities, or specific numeric ranges. Include relevant medical test results with their normal and abnormal ranges. If certain information is not typically applicable or available for this disease, you may omit those sections.

Aim for a comprehensive profile of 30-60 numbered points ensuring that points have quantitative data when applicable.

Focus on the United States.

Follow the structure of this example output:

<example>

1. The prevalence of this condition is estimated to be around 2-3% in the general population.
2. The incidence is more common in females than in males, with a female-to-male ratio of approximately 4-6:1.
3. This condition can manifest at any age, but it is most frequently observed in individuals aged 30-60 years.
4. Risk factors include a positive family history (25-35% increased risk), prolonged exposure to environmental toxins (1.5-2 times increased risk), and dietary deficiencies (15-25% increased risk).
5. The most common cause of this condition is autoimmune inflammation (70% of cases), where the body's immune system mistakenly attacks healthy tissues.
6. The second most common cause of this condition is chronic infection (10-15% of cases), which can trigger a prolonged inflammatory response.
7. The third most common cause of this condition is a genetic mutation (5-10% of cases), leading to abnormal cellular function.

8. The fourth most common cause of this condition is metabolic syndrome (5-7% of cases), where a combination of disorders increases the risk.
9. Less common causes (3-5% of cases) include drug-induced conditions, congenital anomalies, and idiopathic causes.
10. The onset of symptoms is usually gradual but can be sudden in cases triggered by acute stress or infections.
11. The severity of symptoms ranges from mild (grade 10) to severe (grade 70), depending on the extent of the underlying condition and individual variability.
12. Symptoms of this condition may include fatigue (probability 70-80%), joint pain (probability 50-60%), weight changes (probability 40-50%), digestive issues (probability 30-40%), and mood disturbances (probability 50-60%).
13. In autoimmune cases, additional symptoms may include skin rashes (probability 20-30%) and organ-specific dysfunctions (probability 10-20%).

</example>

# Appendix A.2 Prompt for Initial Module Generation

**Full Prompt = Background + Synthea Reference + Synthea Examples + Disease Profile + Generation Prompt**

**Background:**

Synthea is a synthetic patient generator that models the medical history of synthetic patients. Synthea's mission is to create high-quality synthetic, realistic but not real, patient data and associated health records covering every aspect of healthcare. A Synthea module is a JSON document primarily consisting of named states and transitions between the states. The types of states include:
* Control Types: Initial, Terminal, Simple, Guard, Delay, SetAttribute, Counter, CallSubmodule
* Clinical Types: Encounter, EncounterEnd, ConditionOnset, ConditionEnd, AllergyOnset, AllergyEnd, MedicationOrder, MedicationEnd, CarePlanStart, CarePlanEnd, Procedure, ImagingStudy, Device, DeviceEnd, SupplyList. Observation, MultiObservation, DiagnosticReport, Symptom, Death

Synthea modules must strictly follow these rules:
* IMPORTANT! Every state must be the target of a transition from another state, except for the Initial state. There can be no disconnected states.
* Every state except the Terminal state must have an output transition.
* All states much be reachable. This means there must be a directed path from the Initial state to every other state.
* There must be an age guard or delay directly after the initial state. If the disease never happens until a patient is a certain age or age range, use a Delay. If a disease never happens unless a certain condition is met (e.g., the person acquires a certain other condition), use a Guard. If the disease can happen soon after birth use short delay state (for example 1 day, week, or month).
* Every Encounter must be paired with an EncounterEnd.
* Every clinical action (Procedure, Observation, MedicationOrder, ImagingStudy) must occur between an Encounter and EncounterEnd.
* ConditionOnset, ConditionEnd, and Symptom states should not happen during an Encounter. Typically, conditions begin first, symptoms second, and encounters third.
* In distributed transitions, all weights must add up to 100%.
* The name of the state appears as the primary key for the state (as explained in the synthea_module_examples).

Here are some other important things to note:
* Only a percentage of the population will get the disease. Many people never get the disease.
* Not all patients who get the disease get all symptoms. In a Symptom state, the "probability" represents the likelihood of experiencing the symptom, while the range represents the severity of the symptom.
* Symptoms should appear in series, not parallel, unless they are truly mutually exclusive.
* Not all patients get all diagnostic tests. Use your knowledge to determine what tests a patient should receive based on their presentation.

* Use the disease profile to determine which patients get which treatments.
* Every Synthea module must begin with a single Initial state.
* The Terminal state represents the end of the disease course, and the termination of the module. It does not represent death. If patients die, use the Death state.
* Cyclical paths are legal.
* Self-transitions, where the target of the transition is the same state that defines the transition, are legal. A self-transition is used to return to the same state in the next time step and are often used with distributed transitions when there is no progression to the next state.
* Direct transitions transition directly to the indicated state. Distributed transitions transition to one of several possible states based on a probability distribution. For example, a value of 0.55 would indicate a 55% chance of transitioning to the corresponding state. Conditional transitions transition to one of several possible states based on conditional (if-then) logic. A conditional transition consists of an array of condition/transition pairs that are tested in the order they are defined. The first condition that evaluates to true will result in a transition to its corresponding transition state. Conditional logic supports And, Or, Not, At Least, At Most operations upon gender, race, socioeconomic status, and any other patient attribute. Conditional logic can also access the medical record to look at observations, conditions, medications, and care plans.

**Synthea Reference:**

*{A document containing details of the various Synthea states and transitions, abstracted from the Synthea Wiki (https://github.com/synthetichealth/synthea/wiki) sections on States, Logic, and Transitions.}*

**Synthea Examples:**

*{A document containing Synthea module examples, specifically, the validated modules for dementia, food allergies, lupus, sore throat, and urinary tract infections.}*

**Disease Profile:**

*{The specific disease profile produced in Step 1 of the methodology.}*

**Generation Prompt:**

***Task***

Your task is to create a Synthea module for {disease name}.
The disease profile contains requirements for modeling {disease name} that will serve as your basis for building the module.
You are to think step by step about the entire module before generating your final answer.
Each requirement in the disease profile is numbered so it can be easily cited.

***Rules to Follow***

* Every state and transition in the module you create must relate to one or more requirements in the disease profile.
* Every state must include a complete explanation for why it exists. Put the explanation in the "remarks" field.
* Every state must cite a requirement number or numbers from the disease profile that justify the state. Put the reference number(s) in a new property called "requirement number".

***Output***

The output will be MINIFIED JSON with all output compressed into a single line, to minimize the size of the output.

The output will be valid Synthea module following all the rules expressed in the background section.

The "requirement_number" field will contain the fact number or numbers implemented by the state, in this format: "requirement_number": "14, 16, 20"

The "remarks" field will include a complete explanation of the state, its properties, and its transitions.

# Appendix A.3 Prompt for Clinical (Level 2) Review

**Full Prompt = Background + Synthea Reference + Synthea Examples + Disease Profile + Review Prompt + Module to Review**

**Review Prompt:**

***Task***

You are an expert on Synthea, the synthetic patient data simulator. You have mastered the background material and resources provided. Your task is to perform a thorough, thoughtful review of the Synthea module provided.

You will provide a detailed explanation of how each requirement is (or is not) represented in the Synthea module.

Your analysis will be summarized in a numerical score that reflects whether the requirement is adequately modeled in the module.

***Analysis Method***

Your review will use the provided disease profile and confirm that all the requirements in the disease profile are reflected in the Synthea module provided.

You will go through each requirement in the disease profile individually, one at a time, in numerical order, and see if that requirement is correctly implemented.

Points to check:

(a) Is there a state or states in the module that implement the requirement?

(b) Are the properties of that state or those states well-formed and reflective of the requirement?

(c) If the requirement describes a sequence of events, do the states before and after connect to each other in the right order?

(d) Are the input and output transition probabilities and other state parameters consistent with the requirement?

***Output***

The output will be a table with six columns and with one row per requirement:

Column 0 (Requirement_Number) will have the requirement number from the disease profile (1, 2, 3, etc.)

Column 1 (Requirement) will have the full text of the requirement or disease fact quoted from the disease profile. Do not truncate.

Column 2 (Explanation) will have a complete, detailed explanation of how the requirement is (or is not) represented in the Synthea module.

Column 3 (Transitions) will have a complete, detailed explanation of the transitions before, after, and between the states involved with the requirement support the requirement.

Column 4 (Change) will contain a detailed description of changes, if any, that are required to implement the requirement or improve the current implementation. If no change is required, print "none".

Column 5 (Score) will contain a score on how well the requirement has been implemented, as follows:

0.0: The requirement has not been implemented at all or the requirement is not applicable (N/A).

0.25: The requirement has been implemented in some fashion, but the implementation is incorrect.

0.5: The requirement has been partly implemented but does not fully capture the requirement.

0.75: The requirement has been fully implemented but may have minor problem, such as an inaccurate transition probability.

1.0: The requirement has been fully and correctly implemented.

The overall score should account for the requirement, explanation, transitions, and suggested change.

**Module to Review:**

*{Insert the JSON representation of the module under review.}*

## Appendix A.4 Prompt for Iterative Improvement

**Full Prompt = Background + Synthea Reference + Synthea Examples + Update Prompt**

**Update Prompt:**

It looks like you have made some mistakes and/or missed some requirements or just implemented them poorly.
Here is a list of things that can be improved:

*(up to 10 entries follow with values taken from the output of the Level 2 review)*
Requirement *{number}* has not been satisfactorily implemented (score = *{score}*). *{explanation} {transition}*.
RECOMMENDATION: *{recommendation}*.

***Task***

Make the necessary corrections to the Synthea module. Here is the Synthea module you are trying to improve:

*{Insert the JSON representation of the module to be improved.}*

When making modifications to the module, you will not make unnecessary changes,
You will preserve features that are not involved in the current improvements.

***Output***

* The output will be MINIFIED JSON with all output compressed into a single line, to minimize the size of the output.
* The output will be valid Synthea module following all the rules expressed in the background section.
* The "requirement_number" field will contain the fact number or numbers implemented by the state, in this format: "requirement_number": "14, 16, 20"
* The "remarks" field will include a complete explanation of the state, its properties, and its transitions.

# Appendix B: Disease Profiles

## Appendix B.1 Disease Profile Based on Claude Sonnet Intrinsic Knowledge

1. The prevalence of hyperthyroidism is approximately 0.5-1% in the general population.
2. The incidence is higher in females compared to males, with a female-to-male ratio of 5-10:1.
3. Hyperthyroidism can occur at any age, but it is most common in individuals aged 20-50 years.
4. Risk factors include family history (20-30% increased risk), smoking (2-3 times increased risk), and iodine excess (10-20% increased risk).
5. The most common cause of hyperthyroidism is Graves' disease (65% of cases), an autoimmune disorder in which antibodies stimulate the thyroid gland to produce excess hormones.
6. The second most common cause of hyperthyroidism is toxic multinodular goiter (15% of cases), an autonomous function of multiple thyroid nodules, leading to excessive hormone production.
7. The third most common cause of hyperthyroidism is toxic adenoma (10% of cases), a single hyperfunctioning thyroid nodule causing hyperthyroidism.
8. The fourth most common cause of hyperthyroidism is thyroiditis (5% of cases), inflammation of the thyroid gland, causing a transient release of stored hormones.

9. Less common causes (5% of cases) include iodine-induced hyperthyroidism, thyroid hormone resistance, and TSH-secreting pituitary adenomas.
10. The onset of symptoms is usually gradual, but it can be acute in cases of thyroiditis or iodine-induced hyperthyroidism.
11. The severity of symptoms varies from mild (degree 20) to severe (degree 80), depending on the degree of thyroid hormone excess and individual sensitivity.
12. Symptoms of hyperthyroidism can include one or more of the following: palpitations (probability 80-90%), tachycardia (probability 60-70%), atrial fibrillation (probability 5-10%), weight loss despite increased appetite (probability 60-70%), heat intolerance (probability 50-60%), diarrhea (probability 20-30%), nausea (probability 10-20%), tremors (probability 60-70%), muscle weakness (probability 40-50%), hyperreflexia (probability 30-40%), anxiety (probability 50-60%), irritability probability (40-50%), and insomnia (probability 30-40%).
13. In the case of Graves' disease, additional symptoms of hyperthyroidism can include exophthalmos (probability 30-40%), lid lag (probability 20-30%), and diplopia (probability 10-20%).
14. The diagnosis of hyperthyroidism will include a test for Serum TSH. The range 0.1-0.4 mIU/L, indicating low or suppressed TSH, is the hallmark of primary hyperthyroidism (95-100% sensitivity)
15. Tests for elevated Free T4 in the range 1.5-4.5 ng/dL and free T3 in the range 3.0-8.0 pg/mL confirm the diagnosis of hyperthyroidism (90-95% specificity)
16. In the case of suspected Graves' disease, a positive test for TSH receptor antibodies (TRAb) indicates Graves' disease (90-95% sensitivity, 95-100% specificity)
17. Tests for Thyroid peroxidase antibodies (TPOAb) and thyroglobulin antibodies (TgAb), if positive, indicate autoimmune thyroid disorders (50-70% sensitivity, 80-90% specificity)
18. Thyroid ultrasound is used to assess thyroid size, nodularity, and vascularity (80-90% sensitivity)
19. Radioactive iodine uptake (RAIU) is used to differentiate between hyperthyroidism causes (e.g., high uptake in Graves' disease, low uptake in thyroiditis) (80-90% accuracy)
20. Differential diagnoses include non-thyroidal illnesses (e.g., sepsis, heart failure) and drug-induced thyrotoxicosis (e.g., amiodarone, lithium).
21. Treatment with Antithyroid drugs (ATDs), either methimazole or propylthiouracil, are used as first-line therapy in Graves' disease, especially in mild to moderate cases and during pregnancy.
22. Adverse effects of ATDs include agranulocytosis (0.2-0.5%), hepatotoxicity (0.1-0.3%), and skin rash (1-5%).
23. Radioactive iodine (RAI) therapy, the oral administration of I-131, is used as first-line therapy in toxic multinodular goiter and toxic adenoma, and as second-line therapy in Graves' disease.
24. In 80-90% of cases, RAI leads to permanent hypothyroidism, requiring lifelong thyroid hormone replacement.
25. Thyroidectomy, the surgical removal of the thyroid gland, is indicated in large goiters, severe Graves' ophthalmopathy, ATDs and RAI are contraindicated or ineffective
26. Thyroidectomy is successful in 90-100% of cases.
27. Thyroidectomy carries surgical risk of recurrent laryngeal nerve injury and hypoparathyroidis.
28. Thyroidectomy requires lifelong thyroid hormone replacement therapy.
29. Beta-blockers (propranolol, atenolol) are used to control adrenergic symptoms (e.g., palpitations, tremors) until thyroid hormone levels normalize, providing symptomatic relief in 70-80% of cases. Beta-blockers to not treat the underlying hyperthyroidism.
30. Frequent follow-ups (3-6 months) are essential to assess treatment response and adjust therapy as needed.
31. Overall, the mortality risk in treated hyperthyroidism is similar to that of the general population, but untreated or inadequately treated hyperthyroidism may lead to increased cardiovascular morbidity and mortality.
32. Graves' disease remission rates with ATDs range from 30-50% after 12-18 months of therapy.
33. Relapse rates after ATD discontinuation are 50-70%, often requiring definitive therapy with RAI or tyroidectomy.
34. Long-term complications of Graves' disease include Graves' ophthalmopathy (20-30%) and pretibial myxedema (1-2%)
35. For toxic multinodular goiter and toxic adenoma, RAI and surgery provide definitive treatment with a high success rate (90-100%). Recurrence is rare after successful treatment (<5%)
36. Most cases of thyroiditis resolve spontaneously within 2-4 months, but some cases may progress to permanent hypothyroidism (10-20%).

## Appendix B.2 Curated Knowledge Sources for Hyperthyroidism Disease Profile

The following sources were instrumental in creating a comprehensive disease profile for hyperthyroidism. They served as the foundation for the LLM to develop a Synthea module for the condition. These references provided detailed clinical guidelines, epidemiological data, and the latest research insights, ensuring the profile was robust and reflective of current medical knowledge:

## Appendix B.3 Disease Profile With Curated Knowledge and Clinical Expertise

<disease_profile>

<condition>Hyperthyroidism</condition>

<title>Hyperthyroidism Disease Profile</title>

<Background>

- The scope of this module is limited to hyperthyroidism (HT), which happens when the thyroid gland produces and releases excess thyroid hormones, thyroxine and triiodothyronine.
- Thyroid stimulating hormone (also known as thyrotropin), is a pituitary hormone that stimulates the thyroid gland to produce thyroxine and triiodothyronine.
- The three most common causes of HT are: Graves' disease (GD), Toxic Nodular Goiter (TNG), and Thyroiditis (TI).
- Toxic Nodular Goiter (TNG) includes two subtypes, Toxic Adenoma and Toxic Multinodular Goiter.
- Diagnostic tests relevant to HT include total thyroxine (T3), free triiodothyronine (FT4), thyroid stimulating hormone (TSH), thyrotropin receptor antibodies (TRAbs), and radioactive iodine uptake (RAIU).
- Overt HT is defined as low TSH with elevated T3 or elevated FT4 (not necessarily both).
- Subclinical HT is defined as low TSH with normal T3 and FT4.
- Hyperthyroidism risk factors (HRF) include family history of thyroid disease, type 1 or type 2 diabetes, primary adrenal insufficiency, pernicious anemia, and being 0-6 month postpartum.

</Background>

<Assumptions>

- The module will consider GD, TNG, and TI only. Other causes exist but are rarer and are considered out of scope.
- For the purposes of Synthea modeling, Toxic Adenoma and Toxic Multinodular Goiter will not be differentiated and will be considered collectively as TNG.
- HT is rare in children and adolescents under 15, and HT will not be considered for those age groups.
- Except for RAIU, diagnostic tests results are indicated as "low", "normal", or "elevated" rather than quantitatively, since the baseline for these tests changes with age.

- Subclinical HT is assumed to be 40% more prevalent than overt HT across all causes of HT and all age groups.
- The relative incidence of HT causes in the 15 to 60 age group is assumed to be GD: 75%, TNG: 17.5%, TI: 7.5% for both males and females.
- The relative incidence of TNG relative to other causes is assumed to double above age 60, leading to altered distribution of causes: GD: 64%, TNG: 30%, TI: 6% for both males and females.
- Annual incidence of all-cause HT is assumed to be 50% higher in the 60+ age group than the 15-60 age group.
- Lifetime risk of overt HT is assumed to be 3% for women and 0.5% for men.
- The presence of any HRF is assumed to double the rates of HT.
- 15% of patients with HT are assumed to never seek treatment or to be non-compliant with medications, and therefore suffer long-term consequences of HT.

</Assumptions>

<Acronyms>

- ATDs: antithyroid drugs (medication)
- FT4: free T4 (observation)
- GD: Graves' disease (condition)
- HRF: Hyperthyroidism Risk Factor (condition)
- HT: Hyperthyroidism (condition)
- RAI: Radioactive iodine therapy (procedure)
- RAIU: Radioactive iodine uptake (observation)
- RFA: radiofrequency ablation (procedure)
- TI: Thyroiditis (condition)
- TNG: Toxic Nodular Goiter (condition)
- TRAbs: Thyrotropin receptor antibodies (observation)
- TSH: Thyroid stimulating hormone (observation)
- T3: Total Triiodothyronine (observation)

</Acronyms>

<Synthea_Requirements>

<requirement_category>Incidence</requirement_category>

1. Risk of overt GD, women aged 15-60: 1.35%
2. Risk of overt TNG, women aged 15-60: 0.30%
3. Risk of overt TI, women aged 15-60: 0.13%
4. Risk of overt GD, women aged 60+: 0.77%
5. Risk of overt TNG, women aged 60+: 0.36%
6. Risk of overt TI, women aged 60+: 0.07%
7. Risk of overt GD, men aged 15-60: 0.23%
8. Risk of overt TNG, men aged 15-60: 0.05%
9. Risk of overt TI, men aged 15-60: 0.02%
10. Risk of overt GD, men aged 60+: 0.13%
11. Risk of overt TNG, men aged 60+: 0.06%
12. Risk of overt TI, men aged 60+: 0.012%
13. For all population groups, the rate of subclinical HT is 1.4 times the rate of overt HT.
14. The presence of any HRF is assumed to double these rates for all population groups.
15. Of cases of TNG, 80% are mild or moderate and 20% are severe (meaning the patient has large, suspicious, or malignant thyroid nodules).

<requirement_category>Clinical Presentation</requirement_category>

16. Unexplained weight loss despite normal or increased appetite: Affects 70% of HT patients, severity range 20-60%
17. Heart palpitations: Affects 80% of HT cases, severity range 20-70%
18. Insomnia, nervousness, anxiety: Affects 70% of HT cases, severity range 20-80%
19. Heat intolerance and increased sweating: Affects 60% of HT cases, severity range 20-50%
20. Tremors, fatigue, or weakness: Affects 50% of HT cases, severity range 20-80%
21. Hyperdefecation: Affects 25% of HT patients, severity range 20-50%
22. Ophthalmopathy: Affects 33% of HT patients with GD, severity range 20-100%. Severity between 20% and 79% is considered mild to moderate, and above 80% is considered severe.

<requirement_category>Testing and Diagnosis</requirement_category>

23. All patients presenting symptoms of HT should first be tested for T3 and FT4 levels. If subclinical HT is present, then T3 and FT4 tests will be normal. If overt HT is present, tests will reveal elevated T3 and/or elevated FT4.
24. If initial or follow-up testing reveals overt HT, and palpable thyroid nodules are present, or physiologic signs of GD are unclear, then TRAbs should be measured next.
25. However, if initial or follow-up testing reveals overt HT and there are no palpable thyroid nodules and there are clear physiologic signs of GD, TRAbs testing is not necessary, and a diagnosis of GD is confirmed.
26. If TRAbs are elevated, a GD diagnosis is confirmed. TRAbs will be elevated in 98% of cases of GD.
27. Alternatively, if TRAbs are normal, radioactive iodine uptake test (RAIU) should be conducted. However, RAIU is contraindicated in pregnancy and lactation, and a thyroid ultrasound with color-flow Doppler procedure should be substituted.
28. If GD is present, RAIU will reveal diffusely increased uptake in 95% of cases, and then GD diagnosis is confirmed.
29. Alternatively, if TNG is present, RAIU will reveal focal areas of increased uptake, and then TNG diagnosis is confirmed. Nodules revealed by ultrasound likewise indicate the presence of TNG.
30. Alternatively, if TI is present, RAIU will reveal low or absent uptake, and a diagnosis of TI can be confirmed.

<requirement_category>Treatment</requirement_category>

31. If subclinical HT is present (i.e., T3 and FT4 tests are normal), a period of "watchful waiting" commences, during which the T3 and FT4 should be tested at 4-month intervals. For these patients, 3.5% will progress to overt HT per year. (see https://www.ncbi.nlm.nih.gov/pmc/articles/PMC3693616/).
32. Beta-blockers are used for symptomatic relief in 75% of HT patients during initial disease management, particularly those suffering from tachycardia or anxiety.
33. If TI is present, the condition will resolve itself naturally in 1-6 months. Beta blockers can be used for symptomatic relief.
34. If overt GD is present, treatment with antithyroid drugs (ATDs) (e.g., methimazole 5-40 mg/d, propylthiouracil 50-150 mg 3 times daily) is first-line therapy, except in cases of severe TNG or severe ophthalmopathy, in which case surgical intervention is the preferred first-line therapy. After 15 months of treatment with ATDs, there will be remission in 45% of cases, after which, ATDs are discontinued.
35. For patients prescribed ATDs, monitoring of thyroid hormones (T3 and FT4) is done monthly for the first 3 months from the initial treatment.
36. The relapse rate after discontinuation of ATDs is 40% within the first year and 60% within 5 years.
37. If ATD therapy fails to achieve remission after 15 months, or if HT recurs after discontinuation of ATDs, the patient will next receive radioactive iodine therapy (RAI). RAI achieves remission in 85% of cases.
38. Six months after the RAI therapy, the patient's thyroid hormones will be retested. If overt HT persists, the 50% of patients will repeat RAI treatment, and other 50% of patients will move to the third-line therapy, surgical intervention. Patients who have received RAI twice but still have HT will also receive surgical intervention.
39. Surgical intervention is needed for patients with severe TNG, patients with severe ophthalmopathy, and those for whom ATDs and RAI were ineffective. This is 7.5% of overt HT cases overall. Of surgical patients, 80% receive undergo thyroidectomy (procedure) and 20% receive RFA (procedure). Thyroidectomy has a cure rate of 95% and RFA has a success rate of 75%. Patients for whom RFA fails will subsequently undergo thyroidectomy.

40. After successful RAI or surgical intervention (either thyroidectomy or RFA), the risk of hypothyroidism (low thyroid hormone levels) is 85%. For these cases, levothyroxine (medication) is prescribed and continued indefinitely for the life of the patient.
41. Long-term follow-up for overt HT involves annual TSH testing to monitor for recurrence of HT or development of hypothyroidism.

<requirement_category>Untreated and Uncontrolled Patients</requirement_category>

42. 15% of patients with HT have uncontrolled disease, either because they lack access medical care, decline treatment, or are non-compliant with treatments.
43. Long-term osteoporosis risk is increased by 25% in patients with uncontrolled HT.
44. The risk of atrial fibrillation in patients with uncontrolled HT is 12.5%, compared to 1-1.5% in the general population.
45. Long-term risk of cardiovascular mortality is increased by 25% in patients with uncontrolled HT.

</Synthea_Requirements>

</disease_profile>